# Multilayer Collaborative Low-Rank Coding Network for Robust Deep Subspace Discovery

**Xianzhen Li**[1] and **Zhao Zhang**[1,2] and **Yang Wang**[2] and **Guangcan Liu**[3] and **Shuicheng Yan**[4] and **Meng Wang**[2]

**Abstract.** For subspace recovery, most existing low-rank representation (LRR) models performs in the original space in single-layer mode. As such, the deep hierarchical information cannot be learned, which may result in inaccurate recoveries for complex real data. In this paper, we explore the deep multi-subspace recovery problem by designing a multilayer architecture for latent LRR. Technically, we propose a new Multilayer Collaborative Low-Rank Representation Network model termed DeepLRR to discover deep features and deep subspaces. In each layer (>2), DeepLRR bilinearly reconstructs the data matrix by the collaborative representation with low-rank coefficients and projection matrices in the previous layer. The bilinear low-rank reconstruction of previous layer is directly fed into the next layer as the input and low-rank dictionary for representation learning, and is further decomposed into a deep principal feature part, a deep salient feature part and a deep sparse error. As such, the coherence issue can be also resolved due to the low-rank dictionary, and the robustness against noise can also be enhanced in the feature subspace. To recover the sparse errors in layers accurately, a dynamic growing strategy is used, as the noise level will become smaller for the increase of layers. Besides, a neighborhood reconstruction error is also included to encode the locality of deep salient features by deep coefficients adaptively in each layer. Extensive results on public databases show that our DeepLRR outperforms other related models for subspace discovery and clustering.

## 1 INTRODUCTION

Subspace discovery and segmentation via low-rank representation has aroused considerable attention in numerous image processing and computer vision applications [1-10][39-40][44]. To be specific, low-rank subspace discovery aims to segment samples into respective subspaces in the presence of noise or sparse errors [8-11][32-38]. Two most classical low-rank representation models are *Robust Principal Component Analysis* (RPCA) [8] and *Low-Rank Representation* (LRR) [10]. Different from RPCA assuming that samples are drawn from a single low-rank subspace, LRR considers a general problem that samples are from a union of multiple subspaces [9][10]. Specifically, LRR obtains the lowest-rank representation among all the candidates that represent input data as a linear combination of the bases in a dictionary, and the recovered coefficient matrix can be used for clustering subspaces. It is noteworthy that LRR sets the data matrix itself as a dictionary, so if the data sampling is insufficient, the solution of LRR may be trivial if only using the observed data as a dictionary [11]. As such, the recovery result of LRR may not inaccurate in practice. In addition, LRR also cannot process new data directly as it is essentially a transductive model and also it loses important distinguishing features [11].

[1] School of Computer Science and Technology, Soochow University, China, emails: lxz612@outlook.com, cszzhang@gmail.com
[2] Key Laboratory of Knowledge Engineering with Big Data (Ministry of Education) & School of Computer Science and Information Engineering, Hefei University of Technology, Hefei, China, emails: yeungwangresearch @gmail.com, eric.mengwang@gmail.com
[3] Nanjing University of Information Science and Technology, Nanjing, China, email: gcliu@nuist.edu.cn
[4] Department of Electrical and Computer Engineering, National University of Singapore, Singapore, email: eleyans@nus.edu.sg

To solve the insufficient sampling issue and enable LRR for the joint feature extraction, a variant of LRR, called *Latent Low-Rank Representation* (LatLRR) method, was recently proposed. LatLRR recovers the hidden effects by considering the hidden (unobserved) data for defining dictionary [11]. LatLRR clearly unifies the subspace recovery and feature extraction by decomposing data matrix into a principal feature part, a salient feature part and a sparse error part fitting noise. But it is noteworthy that LatLRR cannot preserve the manifold structures of salient features, so it fails to deliver locality preserving features. To solve this issue, a *Laplacian regularized LatLRR* called rLRR [15] was recently proposed to preserve the local geometrical structures of both coefficients and salient features. However, rLRR defines the Laplacian matrix prior to low-rank coding, which is also based on original data, so the recovery results of rLRR may be inaccurate since the pre-obtained Laplacian matrix cannot be ensured to be optimal for the subsequent representations. Moreover, the possibly included noise and corruptions in real data may result in inaccurate similarities. In addition, how to determine the optimal number of the nearest neighbors is unclear and tricky. To solve this issue, recent *Similarity-Adaptive LatLRR* termed SA-LatLRR [16] incorporates a reconstructive error term to correlate the coefficients and salient features in the low-rank coding process, and employs the coefficients matrix as the adaptive reconstruction weights to retain the neighborhood information of salient features. On the other hand, the optimization of LatLRR may be inefficient, because it solves the Nuclear-norm based problem that involves the singular value decomposition (SVD) of the matrices [12], which is usually time-consuming, especially for the large-scale datasets [13][14]. Towards handing this issue, LatLRR based on the Frobenius-norm minimization, termed FLLRR [14], which is as a fast version of LatLRR, has been proposed recently. Specifically, FLLRR directly uses the Frobenius-norm to replace the Nuclear-norm as the surrogate of the rank function [14]. Although FLLRR is more efficient than LatLRR, it still suffers from the other issues mentioned-above as LatLRR.

It is noteworthy that rLRR, SA-LatLRR and FLLRR exhibit attractive properties over LatLRR, but they still suffer from several drawbacks. First, SA-LatLRR, FLLRR, LatLRR and rLRR run the low-rank coding in the original visual space that usually contains various noise and errors to decrease the results. Besides, rLRR and SA-LatLRR are also based on the Nuclear-norm minimization, so their optimizations also suffer from the inefficiency issue. Second, almost all existing LRR methods, including SA-LatLRR, FLLRR, LatLRR and rLRR, use the single-layer architectures. Specifically, they aim at decomposing the input data into several parts only once, i.e., "shallow" representation learning. Due to the fact that the deep hidden semantic and hierarchical information cannot be mined by the single-layer architecture and features are computed based on the original data, the discovered subspaces and learnt feature representation will be inaccurate in reality. Third, existing single layer methods cannot well handle data with high coherence, since the coherence of data plays an important role on subspace learning and it increases as the number of clusters grows. Liu et al. [17] proves that existing low-rank methods can well handle this case that dictionary is low-rank. But existing single-layer methods only consider the self-expressive properties and use the original input data as a dictionary. Thus, it is better to present a new multi-layer low-rank representation model that can learn low-rank dictionaries jointly to solve this issue, which will also be useful to discover and cluster multiple subspaces for complex real data in practice.

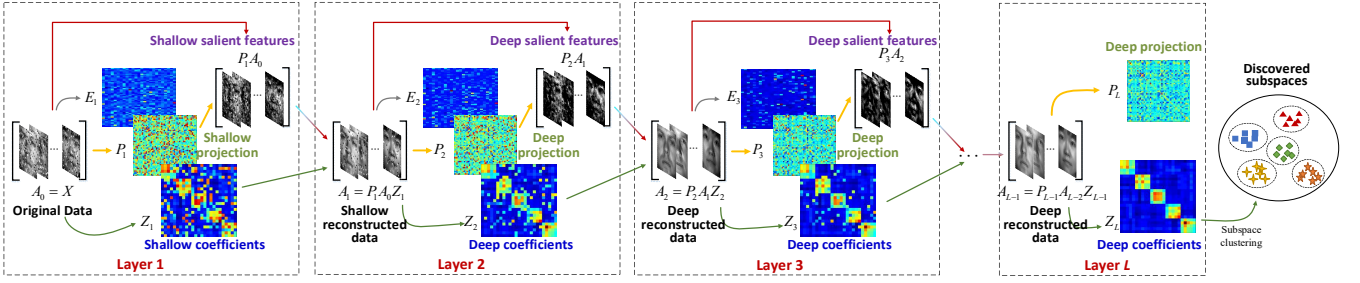

**Figure 1.** The flow-chart of our proposed multilayer collaborative low-rank representation network.

In this paper, we present effective strategies to overcome above drawbacks, and propose a new robust multi-layer low-rank representation model to discover the deep subspace structures and handle the coherence well. The main contributions are shown as:

1. Technically, a novel Multilayer Collaborative Low-Rank Representation Network termed DeepLRR is proposed. DeepLRR clearly extends the low-rank coding from single layer to multi-layers. The advantages of this practice are that multiple layer low-rank structures produce rich and useful deep hidden hierarchical information that has a great potential in learning deep representation and features. To be specific, DeepLRR has the merits of both FLLRR and SA-LatLRR, i.e., computational efficiency and adaptive manifold preservation. But differently, the manifolds are encoded in the discovered deep subspaces of DeepLRR rather that in the original shallow subspaces.

2. To recover multiple-layer low-rank subspaces, DeepLRR uses the collaborative representation scheme to reconstruct the data from row and column directions in each layer (>2), i.e., bilinear reconstruction, which is used as the input of subsequent layers. This operation can fully use the low-rank codes and projection matrices collaboratively from previous layers and can also improve the robust properties of the coding process to noise. The reconstructed deep data of previous layer is directly fed to next layer and is further decomposed into the deep principal features, deep salient features and deep sparse error. Since the noise level of data will be getting less and less with the increase of layers, we use a dynamic growth strategy for the tunable parameter associated with the sparse error when layers are increased. Since DeepLRR adopts the reconstructed low-rank data as the dictionary for low-rank representation rather than the original data, which can clearly address the coherence issue.

3. To encode and preserve the local neighborhoods of deep salient features in the discovered deep subspace, our DeepLRR adds a neighborhood reconstruction error that directly uses the deep coefficients to encode the locality of deep salient features adaptively in each layer, which can also correlate deep salient features using the deep coefficients explicitly in each layer. As a result, the feature representations will be enhanced.

This pager is organized as follows. Section 2 briefly reviews the related work. We propose the formulation of DeepLRR in Section 3. In Section 4, we show the simulation settings and results. Finally, the conclusion is given in Section 5.

## 2 RELATED WORK

### 2.1 Single Layer Low-Rank Representation

**LatLRR.** Given a set of training samples $X = [x_1, x_2 \cdots, x_N] \in \mathbb{R}^{n \times N}$, where $x_i \in \mathbb{R}^n$ is an $n$-dimensional sample and $N$ is sample number, LatLRR solves the insufficient sampling issue by constructing the dictionary by using observed data and hidden data. If the observed and unobserved data are sampled from the same low-rank subspaces, LatLRR recover the hidden effects by minimizing

$$\min_{Z,P,E} \|Z\|_* + \|P\|_* + \lambda \|E\|_1, \quad s.t.\ X = XZ + PX + E, \quad (1)$$

where $\|\cdot\|_*$ is Nuclear-norm [9-12], $\|\cdot\|_1$ is $l_1$-norm for characterizing sparse noise, $XZ$, $PX$ and $E$ are principal features, salient features and sparse error respectively, and $\lambda$ is a positive parameter.

**FLLRR.** LatLRR uses the Nuclear-norm constraints on $Z$ and $P$, so the time-consuming SVD process is involved in each iteration of optimization. Note that Frobenius-norm $\|\cdot\|_F$ can also be used as the convex surrogate of rank function. The replacement is driven by the following relationship between Nuclear-norm, Frobenius-norm and the operator norm $\|\cdot\|$ [14] based on any matrix $A$:

$$\|A\| \leq \|A\|_F \leq \|A\|_* \leq \sqrt{r}\|A\|_F \leq r\|A\|, \quad (2)$$

where $r$ is the rank of $A$ and the operator norm can be seen as the $l_\infty$ norm of $A$. By using the Frobenius-norm to approximate the rank function, the objective function of FLLRR [14] is defined as

$$\min_{Z,P,E} \frac{1}{2}(\|Z\|_F + \|P\|_F) + \lambda \|E\|_1, \quad s.t.\ X = XZ + PX + E. \quad (3)$$

**SA-LatLRR.** Different from LatLRR, SA-LatLRR enhances the representation ability by incorporating a reconstructive error term to correlate the codes and salient features directly, which can also retain local neighborhood information of salient features adaptively. In addition, it regularizes the joint Nuclear-norm and $l_{2,1}$-norm on the feature extracting projection so that extracted salient features are sparse and discriminative. The problem of SA-LatLRR is

$$\min_{Z,P,E} \frac{1}{2}\|Z\|_F^2 + (1-\alpha)\|P\|_* + \alpha\|P\|_{2,1} + \frac{\beta}{2}\|PX - PXZ\|_F^2 + \lambda\|E\|_1, \quad (4)$$
$$\text{s.t. } X = XZ + PX + E,$$

where $\alpha$ and $\beta$ are positive parameters for balancing the terms.

## 2.2 Multilayer Low-rank Representation

With the impressive abilities of representation learning, deep learning based networks achieved a huge success in real-world applications [41-43], mostly by performing supervised learning. However, there are several recent works that discussed the unsupervised deep subspace discovery, e.g., *Deep Subspace Clustering with Sparsity Prior* (PARTY) [25] and *Low-rank Constrained Auto Encoder* (LRAE) [26]. Both PARTY and LRAE used the auto-encoder for subspace clustering. By using the auto-encoder, they incorporate a prior sparsity information (for PARTY) or a prior low-rank information (for LRAE) into the hidden representation layers to keep the sparse or low-rank reconstruction relation based. Since LRAE is clearly based on the low-rank coding, which will be detailed.

**LRAE.** Based on the auto-encoder, LRAE adds more constrains to the network, and minimizes the following objective function:

$$\min_{W_m, b_m} \frac{1}{2}\|X - \hat{X}\|_F^2 + \frac{\lambda_1}{2}\|X_{mid} - X_{mid}Z\|_F^2 + \frac{\lambda_2}{2}\left(\|W_m\|_F^2 + \|b_m\|_F^2\right) \quad (5)$$

for the *m*-th layer of the network, where $W_m$ and $b_m$ denote the weights and bias of the *m*-th layer, $X$ is the input of the network, $\hat{X}$ is output of the network, $X_{mid}$ is the middle representation of the network, $Z$ is the low-rank prior information, $\lambda_1$ and $\lambda_2$ are trade-off parameters. Similar to the typical auto-encoder, LRAE uses the backpropagation algorithm to learn the parameters. When $\lambda_1 = 0$, Eq. (5) degrades to the objective function of auto-encoder [26].

## 3 Proposed Algorithm: DeepLRR

### 3.1 Objective Function

We describe the formulation of DeepLRR in this section. To improve the representation learning, DeepLRR designs a hierarchical and progressive approach, i.e., feeding the reconstructed deep data matrix $P_{l-1}\cdots P_0 XZ_0 \cdots Z_{l-1}$ into the *(l-1)*-th layer into subsequent *l*-th layer, and further decomposing it in the *l*-th layer into a deep low-rank principal feature part $(P_{l-1}\cdots P_0 XZ_0 \cdots Z_{l-1})Z_l$, a deep low-rank salient feature part $P_l(P_{l-1}\cdots P_0 XZ_0 \cdots Z_{l-1})$ and a deep sparse error $E_l$. Assuming that our DeepLRR has *L* layers, the proposed multilayer LRR model can be presented as follows:

$$\begin{aligned}
P_0 XZ_0 &\to (P_0 XZ_0)Z_1 + P_1(P_0 XZ_0) + E_1 \\
P_1 P_0 XZ_0 Z_1 &\to (P_1 P_0 XZ_0 Z_1)Z_2 + P_2(P_1 P_0 XZ_0 Z_1) \\
&\quad + E_2 \\
&\vdots \\
P_{l-1}\cdots P_0 XZ_0 \cdots Z_{l-1} &\to (P_{l-1}\cdots P_0 XZ_0 \cdots Z_{l-1})Z_l \\
&\quad + P_l(P_{l-1}\cdots P_0 XZ_0 \cdots Z_{l-1}) + E_l \\
&\vdots \\
P_{L-1}\cdots P_0 XZ_0 \cdots Z_{L-1} &\to (P_{L-1}\cdots P_0 XZ_0 \cdots Z_{L-1})Z_L \\
&\quad + P_L(P_{L-1}\cdots P_0 XZ_0 \cdots Z_{L-1}) + E_L
\end{aligned} \quad (6)$$

where $Z_l$, $P_l$ and $E_l$ (*l=1,...,L*) are the deep coefficients, deep projection matrix, and deep sparse error matrix of the *l*-th layer, respectively. Note that $P_0$ and $Z_0$ are mainly added to facilitate the descriptions, and are initialized to the identity matrix in our model, i.e., the input of the first layer is original data. It is also should be noted that $Z_{l-1}, ..., Z_0, P_{l-1}, ..., P_0$ are all known variables that are updated in the *(l-1)*-th layer before optimizing $Z_l$, $P_l$ and $E_l$ in the *l*-th layer. The procedure of reconstructing $P_{l-1}\cdots P_0 XZ_0 \cdots Z_{l-1}$ using the deep coefficients and deep projection matrices collaboratively is called collaborative representation in our approach. Intuitively, from the multilayer representation learning process, deep low-rank principal features $(P_{l-1}\cdots P_0 XZ_0 \cdots Z_{l-1})Z_l$ and deep low-rank salient features $P_l(P_{l-1}\cdots P_0 XZ_0 \cdots Z_{l-1})$ are learnt progressively from layers, i.e., extracting fine-grained features from reconstructed deep data matrix from layer to layer. These lead to the following problem for our proposed DeepLRR in the *l*-th layer:

$$\min_{Z_l, P_l, E_l} \frac{1}{2}\left(\|Z_l\|_F^2 + \|P_l\|_F^2\right) + \lambda_l \|E_l\|_1$$
$$\text{s.t. } P_{l-1}\cdots P_0 XZ_0 \cdots Z_{l-1} = (P_{l-1}\cdots P_0 XZ_0 \cdots Z_{l-1})Z_l \quad , \quad (7)$$
$$+ P_l(P_{l-1}\cdots P_0 XZ_0 \cdots Z_{l-1}) + E_l$$

where $\lambda_l$ is a positive tunable parameter that depends on the noise level in data [9][10], i.e., a small $\lambda_l$ is used for a higher level of noise, and else a large $\lambda_l$ is used. Since the noise level of the data matrix will be getting less and less with the increase of layers, we use a dynamic growth strategy for $\lambda_l$ as the layers are increased:

$$\lambda_l = \rho \lambda_{l-1}, \quad l = 1, ..., L, \quad (8)$$

where $\rho \geq 1$ is a control parameter. If $\rho=1$, the same parameter $\lambda_l$ will be used in all the layers.

DeepLRR also considers encoding and preserving the manifold structures of deep salient features in the discovered deep subspace. The neighborhood reconstruction error is formulated as

$$\left\| P_l(P_{l-1}\cdots P_0 XZ_0 \cdots Z_{l-1}) - P_l(P_{l-1}\cdots P_0 XZ_0 \cdots Z_{l-1})Z_l \right\|_F^2. \quad (9)$$

By directly minimizing the neighborhood reconstruction error, one can correlate deep salient features with the deep coefficients $Z_l$ explicitly. Besides, such operation can enable the local structures of deep salient features to be kept in deep subspace, especially in an adaptive manner, because the local neighborhoods in deep features are determined automatically by the deep coefficients $Z_l$.

By combining the neighborhood reconstruction error with Eq.(7), we have the following objective function for DeepLRR:

$$\min_{Z_l, P_l, E_l} \sum_{l=1}^{M} \left\{ \frac{1}{2}\left(\|Z_l\|_F^2 + \|P_l\|_F^2\right) + \frac{\alpha}{2}\|P_l A_{l-1} - P_l A_{l-1} Z_l\|_F^2 + \lambda_l \|E_l\|_1 \right\}, (10)$$
$$\text{s.t. } A_{l-1} = A_{l-1} Z_l + P_l A_{l-1} + E_l, l = 1, ..., L$$

where $A_{l-1} = \tilde{P}_{l-1} X \tilde{Z}_{l-1}$, $\tilde{P}_{l-1} = P_{l-1}\cdots P_0$ and $\tilde{Z}_{l-1} = Z_0 \cdots Z_{l-1}$ are auxiliary matrices included to simplify the formulation. It is clear that the reconstructed deep low-rank data $\tilde{P}_{l-1} X \tilde{Z}_{l-1}$ is also used as dictionary in DeepLRR, i.e., the coherence can be solved to enhance the subspace learning in the case of discovering multiple subspaces.

### 3.2 Optimization

We show the optimization of DeepLRR. To deliver the enhanced representations, we train the model layer by layer. From the objective function, DeepLRR with *L* layers can be divided into *L* sub-problems. For the *l*-th layer, the target function is defined as

$$\min_{Z_l, P_l, E_l} \frac{1}{2}\left(\|Z_l\|_F^2 + \|P_l\|_F^2\right) + \frac{\alpha}{2}\|P_l A_{l-1} - P_l A_{l-1} Z_l\|_F^2 + \lambda_l \|E_l\|_1, \quad (11)$$
$$\text{s.t. } A_{l-1} = A_{l-1} Z_l + P_l A_{l-1} + E_l$$

**Algorithm 1:** Solving Eq. (10) by Inexact ALM (*l*-th layer)
**Inputs:** Reconstructed data $A_{l-1}$, tunable parameters $\lambda_l$, $\alpha_l$.
**Initialization:** $Z_l^0 = 0$, $P_l^0 = 0$, $E_l^0 = 0$, $Y^0 = 0$, $\mu_{max} = 10^6$, $\mu^0 = 10^{-6}$, $\eta = 1.5$, $\varepsilon = 10^{-7}$, $k = 0$.
**While** *not converged* **do**
1. Fix the others and update the coefficients $Z_l^{k+1}$ by Eq.(14);
2. Fix the others and update the projection matrix $P_l^{k+1}$ by Eq.(16);
3. Fix the others and update the sparse error $E_l^{k+1}$ by Eq.(17);
4. Update the Lagrange multiplier $Y^{k+1}$;
5. Update the parameter $\mu$ by $\mu^{k+1} = \min(\eta\mu^k, \mu_{max})$;
6. Check for convergence: If $\|A_{l-1} - A_{l-1}Z_l^{k+1} - P_l^{k+1}A_{l-1} - E_l\|_\infty \leq \varepsilon$, stop; else $k = k+1$ and return to step 1.
**End while**
**Output:** $Z_l^* = Z_l^{k+1}$, $P_l^* = P_l^{k+1}$.

---

**Algorithm 2:** Optimization procedures of DeepLRR
**Input:** Observation data matrix $X$, layer number $L$, tunable parameters $\alpha_l$ and $\lambda_1$, and positive growing step $\rho$. Initialize $Z_0$ and $P_0$ to be the identity matrices.
**For** *l*=1 to *L*, **do**
1. Compute the reconstructed data matrix $A_{l-1}$ as
$$A_{l-1} = P_{l-1} \cdots P_0 X Z_0 \cdots Z_{l-1};$$
2. Update the tunable parameter $\lambda_l$ by $\lambda_l = \rho\lambda_{l-1}(l \geq 2)$;
3. Solve $Z_l$, $P_l$ and $E_l$ by optimizing Eq. (10) in Algorithm 1.
**End**
**Output:** $Z_l$ and $P_l$, where *l*=1, 2,…, *L*.

---

where $A_{l-1} = \tilde{P}_{l-1} X \tilde{Z}_{l-1}$. It is easy to check that $Z_l, P_l, E_l$ depend on each other, so they cannot be solved directly. Following common procedures, we solve Eq. (11) by updating $Z_l, P_l, E_l$ alternately, i.e., solving one of them by fixing others. For efficiency, we use the inexact Augmented Lagrange Multiplier (Inexact ALM) method [12]. The Lagrange function of Eq. (11) can be constructed as

$$\wp = \frac{1}{2}\left(\|Z_l\|_F^2 + \|P_l\|_F^2\right) + \frac{\alpha}{2}\|P_l A_{l-1} - P_l A_{l-1} Z_l\|_F^2 + \lambda_l \|E_l\|_1$$
$$+ tr\left(Y^T\left(A_{l-1} - A_{l-1} Z_l - P_l A_{l-1} - E_l\right)\right), \quad (12)$$
$$+ \frac{\mu}{2}\|A_{l-1} - A_{l-1} Z_l - P_l A_{l-1} - E_l\|_F^2$$

where $Y$ is Lagrange multiplier and $\mu$ is a positive factor. By using inexact ALM, the optimization of DeepLRR can be detailed as

**(1) Fix others, update $Z_l$ in the *l*-th layer:**

In the *l*-th layer, $Z_{l-1},\ldots,Z_0, P_{l-1},\ldots,P_0$ are all known variables that are updated in the *(l-1)*-th layer. With other variables fixed, we can update the $Z_l$ from the following reduced problem:

$$J(Z_l) = \frac{1}{2}\|Z_l\|_F^2 + \frac{1}{2}\left\|\begin{pmatrix}\sqrt{\alpha}P_l A_{l-1} \\ \sqrt{\mu}\Xi_l\end{pmatrix} - \begin{pmatrix}\sqrt{\alpha}P_l A_{l-1} \\ \sqrt{\mu}A_{l-1}\end{pmatrix}Z_l\right\|_F^2, \quad (13)$$
$$+ tr\left(Y^T(\Xi_l - A_{l-1}Z_l)\right)$$

where $\Xi_l = A_{l-1} - P_l A_{l-1} - E_l$. Denote $\Lambda_l = \left(\sqrt{\alpha}(P_l A_{l-1})^T, \sqrt{\mu}\Xi_l^T\right)^T$, $\Delta_l = \left(\sqrt{\alpha}(P_l A_{l-1})^T, \sqrt{\mu}A_{l-1}^T\right)^T$, by taking the derivative of $J(Z_l)$ w.r.t. $Z_l$ and zeroing the derivative, we can infer the coding coefficients matrix $Z_l$ at the *(k+1)*-th iteration as follows:

$$Z_l^{k+1} = \left(I + \Delta_l^{kT}\Delta_l^k\right)^{-1}\left(\Delta_l^{kT}\Lambda_l^k + A_{l-1}^T Y^k\right), \quad (14)$$

where $Y_l^k$ is the a Lagrange multiplier, $I$ is the identity matrix, matrices $\Xi_l^k = A_{l-1} - P_l^k A_{l-1} - E_l^k$, $\Lambda_l^k = \left(\sqrt{\alpha}(P_l^k A_{l-1})^T, \sqrt{\mu^k}\Xi_l^{kT}\right)^T$, and $\Delta_l^k = \left(\sqrt{\alpha}(P_l^k A_{l-1})^T, \sqrt{\mu^k}A_{l-1}^T\right)^T$ in the *l*-th layer.

**(2) Fix others, update $P_l$ in the *l*-th layer:**

We then discuss the optimization of $P_l$ in the *l*-th layer. In this step, other variables are constants. We can then update $P_l$ from

$$J(P_l) = \frac{1}{2}\|P_l\|_F^2 + \frac{\alpha}{2}\|P_l\Phi_l\|_F^2 + tr\left(Y^T(\Phi_l - E_l - P_l A_{l-1})\right), \quad (15)$$
$$+ \frac{\mu}{2}\|\Phi_l - E_l - P_l A_{l-1}\|_F^2$$

where $\Phi_l = A_{l-1} - A_{l-1}Z_l$ is an auxiliary matrix. By taking the derivative of $J(P_l)$ w.r.t. $P_l$ and zeroing it, we can infer the projection matrix $P_l$ at the *(k+1)*-th iteration as follows:

$$P_l = \left(Y^k A_{l-1}^T + \mu^k\left(\Phi_l^k - E_l^k\right)A_{l-1}^T\right)\left(I + \alpha\Phi_l^k\Phi_l^{kT} + \mu^k A_{l-1}^T A_{l-1}\right)^{-1}, \quad (16)$$

where $\Phi_l^k = A_{l-1} - A_{l-1}Z_l^{k+1}$ is the auxiliary matrix in the *l*-th layer.

**(3) Fix others, update $E_l$ in the *l*-th layer:**

Finally, we present the optimization of $E_l$. In this step, other variables are fixed and are regarded as constants. By removing the terms irrelevant to $E_l$ and making some easy transformation, we have

$$\min_{E_l} \frac{\lambda_l}{\mu^k}\|E_l\|_1 + \frac{1}{2}\left\|E_l - \left(A_{l-1} - A_{l-1}Z_l^{k+1} - P_l^{k+1}A_{l-1} + Y^k/\mu^k\right)\right\|_F^2, \quad (17)$$

from which the iterate $E_l^{k+1}$ can be obtained by the shrinkage operator as $E_l^{k+1} = \Omega_{\lambda_l/\mu^k}[\Sigma_E]$ [12], where the shrinkage operator is defined as $\Omega_\varepsilon[x] = sgn(x)\max(|x| - \varepsilon, 0)$ and the matrix $\Sigma_E$ is defined as $\Sigma_E = A_{l-1} - A_{l-1}Z_l^{k+1} - P_l^{k+1}A_{l-1} + Y^k/\mu^k$.

For complete presentations of our approach, we summarize the procedures of solving the problem of Eq. (10) in Algorithm 1.

### 3.3 Summary of DeepLRR Algorithm

Since the objective function of our DeepLRR with $L$ layers can be divided into $L$ similar sub-problems as Eq. (10), we can obtain $L$ coefficient matrices $Z_1^*,\ldots,Z_L^*$ and $L$ projection matrices $P_1^*,\ldots,P_L^*$, where $Z_L^*$ and $P_L^*$ are called the deepest coefficient and projection matrices. Then, we can compute the optimal deep low-rank salient features as $P_L^* P_{L-1}^* \cdots P_1^* X Z_1^* \cdots Z_{L-1}^*$ and deep low-rank principal features as $P_{L-1}^* \cdots P_1^* X Z_1^* \cdots Z_{L-1}^* Z_L^*$. Besides, the deep reconstructed data after $L$ layers can be obtained as $P_L^* P_{L-1}^* \cdots P_1^* X Z_1^* \cdots Z_{L-1}^* Z_L^*$. For complete presentation, we summarize the procedures in Algorithm 2.

### 3.4 Discussion

We discuss the relations of DeepLRR to SA-LatLRR and FLLRR in single-layer case. Specifically, when $L$=1, the objective function of our DeepLRR can be reduced to

$$\min_{Z_1, P_1, E_1} \frac{1}{2}\left(\|Z_1\|_F^2 + \|P_1\|_F^2\right) + \frac{\alpha}{2}\|P_1 X - P_1 X Z_l\|_F^2 + \lambda_1 \|E_1\|_1, \quad (18)$$
$$s.t.\ P_0 X Z_0 = P_0 X Z_0 Z_1 + P_1 P_0 X Z_0 + E_1$$

which is just the Frobenius-norm based objective function of SA-LatLRR, since $P_0$ and $Z_0$ are identity matrices. If we further constrain $\alpha$=0, the reduced problem identifies the objective function of FLLRR. Thus, FLLRR and SA-LatLRR are the special examples of DeepLRR in single-layer case. As such, DeepLRR will be superior to FLLRR and SA-LatLRR for recovering the subspaces.

### 3.5 Computational Time Complexity

We analyze the time complexity of each layer in Algorithm 1. For DeepLRR, no SVD is involved and the major computation is from the matrix inversions in Step 1 and Step 2. So, the time complexity of Algorithm 1 is reduced to 1/2 of that of LatLRR, which is equal to that of FLLRR. Thus, it is easy to infer that the total time complexity of DeepLRR in Algorithm 2 is $L$ times that of Algorithm 1, where $L$ is the number of layers and is usually a small value.

## 4 Experimental Result and Analysis

We perform simulations to show the effectiveness of our DeepLRR. The other compared algorithms includes LRR [9][10], LatLRR [11], FLLRR [14], RPCA+LRR [17], BDR [23], rLRR [15], SA-LatLRR [16] and LRAE [26]. For fair comparison, the setting of LRAE is set the same as [26] and all parameters of each compared method are chosen carefully. All experiments are performed on a PC with Intel(R) Core(TM) i7-3770 CPU @ 3.40GHz.

### 4.1 Data Preparation

Four popular real databases, including face and object image databases, are evaluated. Face databases include Extended Yale B [20], UMIST [21] and AR [24]. Object database includes COIL100 [22]. Fig.2 shows some image examples of used datasets. As a common practice, all the images are down-sampled, i.e., resize the images of Yale B into $32 \times 28$ pixels, resize the images of other databases into $32 \times 32$ pixels. The pixel values are also normalized into [0, 1].

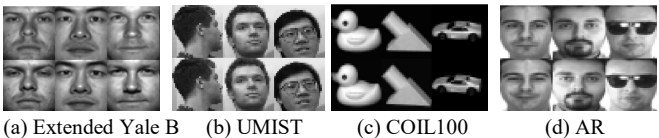

(a) Extended Yale B (b) UMIST (c) COIL100 (d) AR
**Figure 2.** Image examples of evaluated real image databases.

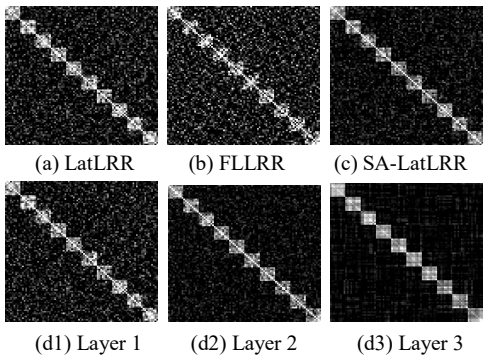

(a) LatLRR (b) FLLRR (c) SA-LatLRR
(d1) Layer 1 (d2) Layer 2 (d3) Layer 3
**Figure 3.** Visualization of the coefficient matrices of LatLRR (a), FLLRR (b), SA-LatLRR (c) and DeepLRR (d1-d3) (Layer 1, Layer 2, Layer 3).

### 4.2 Visual Image Analysis by Visualization

**Visualization of coefficients matrix.** We first visualize the coefficients matrix of each algorithm for comparison. For a coefficient matrix with desired block-diagonal structures, each block includes the codes for certain class so that each sample can be reconstructed by the samples of one class as much as possible. An artificial dataset is sampled and evaluated. We follow the procedures in [16] to construct 10 independent subspaces $\{S_i\}_{i=1}^{10}$ whose bases $\{B_i\}_{i=1}^{10}$ are calculated by $B_{i+1}=RB_i$, $i=1, 2, ..,9$, where $R$ is a random rotation matrix and $B_i$ is a random column orthogonal matrix whose dimensionality is $200 \times 10$, so the rank of each subspace is 10. Then, an artificial data matrix $X=[X_1, X_2,..., X_{10}] \in R^{200 \times 90}$ is formed by sampling 9 data vectors from each subspace by $X_i=B_iC_i$, $i=1, 2, ..., 9$, where $C_i$ is an i.i.d. $N(0,1)$ matrix whose dimension is $10 \times 9$. The random Gaussian noise with variance 0.1 in added into the samples to evaluate the robust properties of each method. Fig.3 visualizes the coefficient matrices of the latent approaches LatLRR, FLLRR, SA-LatLRR, and DeepLRR. For our DeepLRR, we visualize the coefficient matrices of the first three layers. From the results, we can easily see that: (1) DeepLRR obtains clear block-diagonal structures compared with other methods, when layers become deeper; (2) the coefficient matrices from the 2 and 3 layers of our DeepLRR contain less wrong inter-class connections and better connectivity, which can produce more accurate subspace representation potentially, compared with that of the first layer.

**Visualization of principal features.** We visualize the principal features or reconstructed data by the coefficients. Extended Yale B database is evaluated as an example, and we randomly select 10 samples each category for representation learning. Then, 20% pixel corruptions are included into training set. The visualization results of recovered principal features are shown in Fig. 4, where we also show the features from the first 3 layers of our DeepLRR for comparison. We see that the recovered principal features of DeepLRR are more accurate with the increasing number of layers.

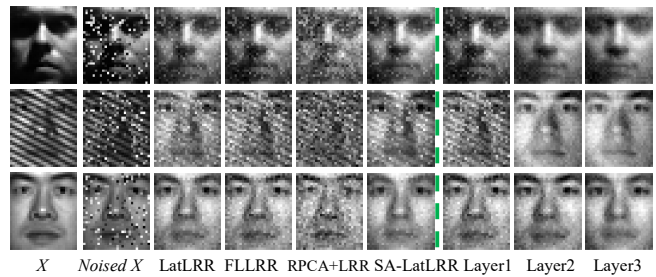

X   Noised X   LatLRR  FLLRR  RPCA+LRR  SA-LatLRR  Layer1  Layer2  Layer3
**Figure 4.** Visualization of recovered principal features by each method.

**Visualization of the decomposition process of DeepLRR.** We visualize the latent representation learning process of our method. Specifically, for a given data matrix $X$, our DeepLRR decomposes it into a deep low-rank principal feature part $(P_{l-1} \cdots P_0 X Z_0 \cdots Z_{l-1})Z_l$, a deep low-rank salient feature part $P_l(P_{l-1} \cdots P_0 X Z_0 \cdots Z_{l-1})$ and a deep sparse error $E_l$, where $l$ is the number of layers. We visualize the recovered deep low-rank principal and salient features in Fig. 5, where we also show the results in the first 3 layers for comparison. We can observe that the deep principal features correspond to the average faces, and the deep salient features correspond to the distinguishing key parts of faces, such as noses and eyes. It can also be found that the occlusion and shadow in the face images can be recovered clearly along with the increasing number of layers.

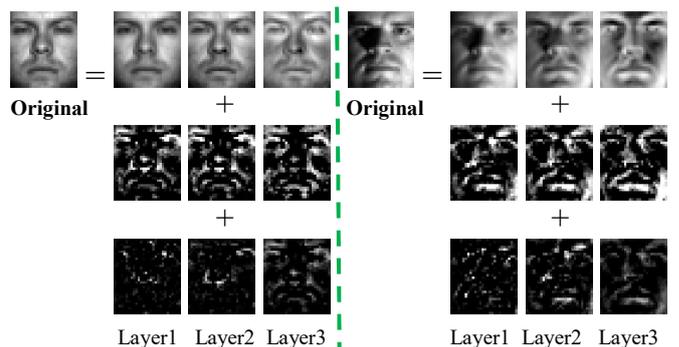

Layer1 Layer2 Layer3    Layer1 Layer2 Layer3
**Figure 5.** Visualization of principal features *(first row)*, salient features *(second row)* and errors *(third row)* of three image examples by DeepLRR.

### 4.3 Application to Image Clustering

In this experiment, we evaluate the subspace clustering power of DeepLRR and other related algorithms on several public databases,

including Extended Yale B, UMIST, COIL100 and AR. For the Extended Yale B, COIL100 and AR, we randomly select $S$ images from each category for evaluations, where $S$ is chosen from {5, 10, 15, 20}. For UMIST, we randomly select $C$ categories for evaluation, where $C$ is chosen from {10, 15, 20}.

**Evaluation procedure.** To evaluate the clustering performance, we use the coefficient matrix $Z^*$ of each method to define the edge weights of an undirected graph, i.e., $W = (|Z^*|+|Z^{*T}|)/2$, and then use the NCut method to produce the data clustering result. For our DeepLRR, we use the coefficient matrix $Z_L^*$ obtained from the $L$-th layer to define the edge weights as $W = (|Z_L^*|+|Z_L^{*T}|)/2$.

**Evaluation metric.** We apply three widely-used popular metrics, i.e., clustering accuracy (ACC), normalized mutual information (NMI) [19] and F-score. Specifically, the larger the values of ACC, NMI and F-score, the better the clustering results.

**Data clustering results.** We show the clustering result based on each database in Table 1. The final clustering result of each method is averaged based on 100 runs to avoid the randomness. From the results, we see that: (1) DeepLRR delivers higher clustering results than other related methods in most cases, especially for AR whose images are partly covered by glasses or scarf. The results means that DeepLRR can perform the subspace segmentation more accurately; (2) the clustering result of each model usually goes down as the number of categories or samples increases, since clustering the dataset with less number of categories or samples is easier than more. FLLRR is highly comparable to LatLRR, which means that the Frobenius-norm indeed can be used as the surrogate of the rank function in reality; (3) compared with the deep auto-encoder based LRAE that learns the deep low-rank features for subspace clustering, DeepLRR can perform better in most cases.

### 4.4 Clustering Images against Corruptions

We evaluate the robustness of each method for clustering images with pixel corruptions. Extended Yale B and UMIST are involved in this study. For Extended Yale B, we choose 10 images per category, and then manually corrupt data with the Gaussian noise under different variances, i.e., 0, 50, 100, 150, 200, 250 and 300 respectively. Fig.6 shows the clustering results on Yale B, and also illustrates some examples of noisy images. For UMIST, we randomly select images of 10 categories and manually corrupt a percentage of selected pixels by replacing the gray values with random corrupted values from [0,1]. The percentage of corruptions varies from 0% to 60%, and the results on UMIST is shown in Fig.7.

We have the following observations: 1) from Fig. 6, the clustering results decrease with the increasing variance, since the noise intensity has increasing negative effects on representations; 2) from Fig.7, the clustering results keep stable or decrease slowly for the range of 0%~50% for most methods; 3) DeepLRR delivers better results than other methods for robust learning against corruptions.

### 4.5 An Ablation Study

We show the parameter sensitivity analyses of DeepLRR. Since the parameter selection issue is still an open issue, a heuristic way is to select the most important ones. We mainly explore the effects of parameters of DeepLRR on the results, i.e., trade-off parameter $\lambda_1$ of the first layer, step size $\rho$ and the parameter $\alpha$ in each layer.

**Selection of parameters $\lambda_1$ and $\alpha$.** We randomly select 10 samples from each category of COIL100, and we select 20 categories of UMIST randomly. Then, we tune the parameters $\lambda_1$ and $\alpha$ by grid search from the candidate set [$10^{-8}$, $10^{-6}$, ..., $10^6$, $10^8$], and the results are shown in the left 3D histogram of Fig.8 (a-b), respectively. We see that our proposed DeepLRR is very robust to the model parameter $\alpha$ for fixed $\lambda_1$, and higher clustering results are obtained with the range of $10^{-8} \leq \lambda_1 \leq 10^{-2}$.

**Selection of step size $\rho$ and number of layers ($L$).** With $\alpha$ and $\lambda_1$ obtained, we can explore tuning $\rho$ and $L$ by the grid search from the candidate sets [5, 10, 50, 100, 500, 1000] and [$10^1$, $10^2$, ..., $10^6$], respectively. The parameter selection results of $\rho$ and $L$ are shown in right of Fig.8 (a-b). We see that the clustering performance of DeepLRR can be improved for the range of 2 to 5 layers, and the best records are obtained at $L=3$ or 5 in the investigated cases. Since the step size $\rho$ depends on the level of noise in each layer, its choices are slightly different.

### 4.6 Convergence Analysis

Since the objective function of DeepLRR is optimized from layer to layer, in this study we mainly present some convergence analysis results. COIL100 object and AR face databases are involved in this study. We randomly select 5 samples from each category for COIL100 to the low-rank representation learning. We present the averaged convergence results of first 5 layers of DeepLRR in Fig.9. From the results, we find that: (1) DeepLRR converges rapidly in each layer, and specifically it converges with the number of iterations ranging from 30 to 80 in most cases; (2) With the increasing number of iterations, both the running time and the rank of reconstructed data decreases gradually, which means that our DeepLRR spends less time when the number of layers is increased.

## 5 DISCUSSION AND FUTURE WORK

We have discussed the robust deep subspace discovery, and technically proposed a simple yet effective subspace recovery and learning model called Multilayer Collaborative Low-Rank Representation Network. DeepLRR learns the deep low-rank representation by designing a multi-layer architecture, which can handle the complex data well in a progressive way. DeepLRR extends the regular low-rank coding from single layer to multiple layers, which can generate useful hierarchical information that is great potential for learning representations. For the representation learning, DeepLRR aims to reconstruct data from both row and column directions in each layer, which is set as the input of subsequent layers by integrating the hierarchical information. We also include a neighborhood reconstruction error based on deep salient features and deep coefficients to encode the locality of salient features in each layer.

We evaluated DeepLRR by several public image datasets. The clustering results on original and corrupted images show the superior performance of our method compared with several closely-related models. The visual image analysis demonstrates that the obtained coefficients matrix has clear block-diagonal structures. We have also proposed an effective heuristic strategy to select the hyper-parameters of DeepLRR and testified its validity. However, the optimal selection of model parameters is still challenging. In future, we will explore how to incorporate labeled and unlabeled data [45-46] for deep semi-supervised low-rank representation.

## ACKNOWLEDGEMENTS

This work is partially supported by the National Natural Science Foundation of China (61672365, 61732008, 61725203, 61622305, 61871444 and 61806035) and the Fundamental Research Funds for the Central Universities of China (JZ2019- HGPA0102). Dr. Zhao Zhang is the corresponding author.

Table 1. Numerical comparison of clustering results on the four real-world image databases.

| | | | LRR | LatLRR | FLLRR | RPCA+LRR | BDR-B | BDR-Z | rLRR | SA-LatLRR | LRAE | **DeepLRR** |
|---|---|---|---|---|---|---|---|---|---|---|---|---|
| Extended Yale B | ACC | S=10 | 0.7044 | 0.6899 | 0.7099 | 0.7416 | 0.6945 | 0.6902 | 0.5838 | 0.7408 | 0.5395 | **0.7463** |
| | | S=15 | 0.6476 | 0.6716 | 0.6785 | 0.7109 | 0.6697 | 0.6721 | 0.5308 | **0.7123** | 0.5314 | 0.6940 |
| | | S=20 | 0.6037 | 0.6157 | 0.6042 | 0.6393 | 0.6215 | 0.6102 | 0.5026 | 0.6294 | 0.4723 | **0.6403** |
| | NMI | S=10 | 0.8001 | 0.8289 | 0.8255 | 0.8415 | 0.8005 | 0.7946 | 0.7384 | 0.8434 | 0.7187 | **0.8572** |
| | | S=15 | 0.7312 | 0.7752 | 0.7797 | 0.8035 | 0.7651 | 0.7637 | 0.6721 | 0.8235 | 0.6969 | **0.8385** |
| | | S=20 | 0.6806 | 0.7154 | 0.7061 | 0.7390 | 0.7047 | 0.6858 | 0.6243 | 0.7172 | 0.6369 | **0.7484** |
| | F-Score | S=10 | 0.4566 | 0.4473 | 0.5275 | 0.4714 | 0.4423 | 0.4086 | 0.2867 | 0.4579 | 0.3840 | **0.5615** |
| | | S=15 | 0.3943 | 0.4170 | 0.4311 | 0.3439 | 0.3759 | 0.3537 | 0.2365 | 0.4617 | 0.3951 | **0.4802** |
| | | S=20 | 0.3456 | 0.3762 | 0.3495 | 0.2842 | 0.2853 | 0.2521 | 0.2205 | **0.3941** | 0.3333 | 0.3680 |
| UMIST | ACC | C=10 | 0.6902 | 0.7288 | 0.7216 | 0.6903 | 0.7235 | 0.7048 | 0.604 | 0.7381 | 0.6788 | **0.7687** |
| | | C=15 | 0.6469 | 0.5584 | 0.6186 | 0.661 | 0.6377 | 0.6415 | 0.6225 | 0.6884 | 0.6224 | **0.6908** |
| | | C=20 | 0.5236 | 0.4758 | 0.5585 | 0.5684 | 0.5249 | 0.538 | 0.4778 | 0.5541 | 0.5678 | **0.6192** |
| | NMI | C=10 | 0.7495 | 0.7641 | 0.7613 | 0.7519 | 0.7881 | 0.7883 | 0.7508 | 0.7807 | 0.7796 | **0.8158** |
| | | C=15 | 0.7528 | 0.7073 | 0.7352 | 0.7540 | 0.7436 | 0.7478 | 0.7429 | **0.7809** | 0.7665 | 0.7676 |
| | | C=20 | 0.7058 | 0.6803 | 0.7135 | 0.7204 | 0.7047 | 0.6957 | 0.7042 | 0.6935 | **0.7650** | 0.7400 |
| | F-Score | C=10 | 0.6371 | 0.6341 | 0.6375 | 0.6389 | 0.6791 | 0.6630 | 0.5947 | 0.6660 | 0.6543 | **0.725** |
| | | C=15 | 0.6078 | 0.4056 | 0.5563 | 0.5941 | 0.5715 | 0.5801 | 0.5627 | 0.6202 | 0.5838 | **0.6217** |
| | | C=20 | 0.4947 | 0.3448 | 0.4881 | 0.4917 | 0.4472 | 0.4585 | 0.3993 | 0.4886 | **0.5558** | 0.5501 |
| COIL100 | ACC | S=5 | 0.8215 | 0.7942 | 0.8298 | **0.8809** | 0.8086 | 0.8457 | 0.7019 | 0.8090 | 0.7702 | 0.8662 |
| | | S=10 | 0.7074 | 0.6985 | 0.7231 | 0.7370 | 0.6533 | 0.6539 | 0.6361 | 0.7059 | 0.7306 | **0.7486** |
| | | S=15 | 0.5921 | 0.6003 | 0.6263 | 0.6357 | 0.5519 | 0.5476 | 0.5778 | 0.637 | **0.6497** | 0.6484 |
| | NMI | S=5 | 0.9588 | 0.9481 | 0.9597 | **0.9695** | 0.9577 | 0.9646 | 0.9183 | 0.9552 | 0.9522 | 0.9689 |
| | | S=10 | 0.9188 | 0.9156 | 0.9233 | 0.9209 | 0.8981 | 0.8989 | 0.8885 | 0.9192 | 0.9216 | **0.9249** |
| | | S=15 | 0.8488 | 0.8643 | 0.8692 | 0.8737 | 0.8211 | 0.8083 | 0.8476 | 0.8739 | **0.8810** | 0.8755 |
| | F-Score | S=5 | 0.7467 | 0.6881 | 0.7467 | 0.8094 | 0.7102 | 0.7552 | 0.3979 | 0.7354 | 0.7363 | **0.8257** |
| | | S=10 | 0.6422 | 0.6527 | 0.6778 | 0.6836 | 0.5109 | 0.5029 | 0.3074 | 0.6567 | 0.6844 | **0.6997** |
| | | S=15 | 0.4430 | 0.5465 | 0.5593 | 0.565 | 0.2551 | 0.2073 | 0.5137 | 0.5632 | **0.5893** | 0.5866 |
| AR | ACC | S=10 | 0.6104 | 0.5686 | 0.6206 | 0.6283 | 0.4856 | 0.4556 | 0.4948 | 0.5380 | 0.5079 | **0.6621** |
| | | S=15 | 0.6815 | 0.6354 | 0.6929 | 0.7089 | 0.6206 | 0.6239 | 0.4941 | 0.6066 | 0.5422 | **0.7716** |
| | | S=20 | 0.6564 | 0.6000 | 0.6715 | 0.6898 | 0.6210 | 0.6304 | 0.4841 | 0.5817 | 0.4818 | **0.7536** |
| | NMI | S=15 | 0.8155 | 0.7943 | 0.8277 | 0.8310 | 0.7548 | 0.6910 | 0.726 | 0.7793 | 0.7892 | **0.8472** |
| | | S=20 | 0.8297 | 0.8028 | 0.8419 | 0.8512 | 0.7966 | 0.7964 | 0.7137 | 0.7813 | 0.7814 | **0.8984** |
| | | S=10 | 0.8046 | 0.7741 | 0.8211 | 0.8295 | 0.7818 | 0.7853 | 0.7017 | 0.7642 | 0.7157 | **0.8774** |
| | F-Score | S=15 | 0.4234 | 0.3743 | 0.4456 | 0.4258 | 0.2864 | 0.2225 | 0.303 | 0.3353 | 0.3423 | **0.4649** |
| | | S=20 | 0.4609 | 0.4139 | 0.4489 | 0.4908 | 0.346 | 0.3307 | 0.316 | 0.3951 | 0.3972 | **0.5978** |
| | | S=10 | 0.4435 | 0.3947 | 0.4698 | 0.4253 | 0.3719 | 0.3801 | 0.3212 | 0.3763 | 0.3332 | **0.5288** |

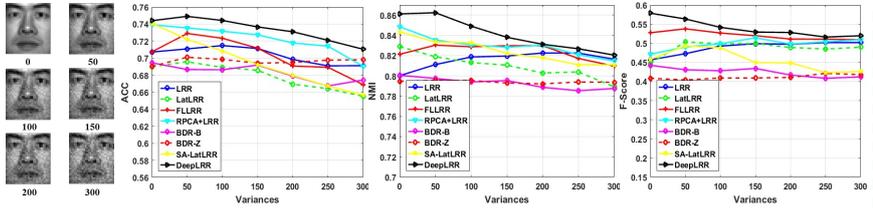

Figure 6. Clustering results of each algorithm vs. varying variance on Extended Yale B.

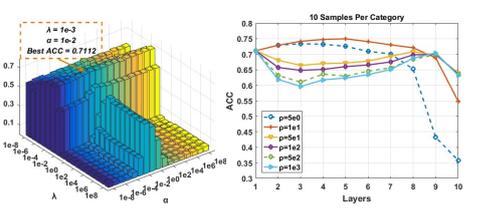

Figure 8 (a). 10 samples per category in COIL100.

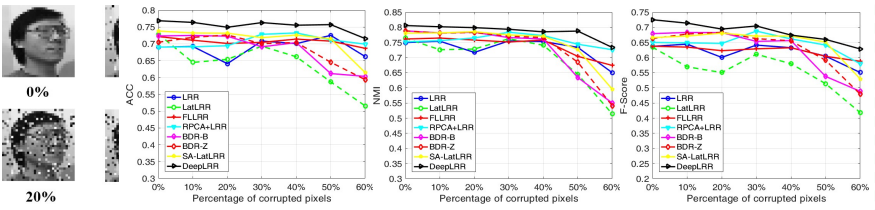

Figure 7. Clustering results of each algorithm vs. varying percentage of corrupted pixels on UMIST.

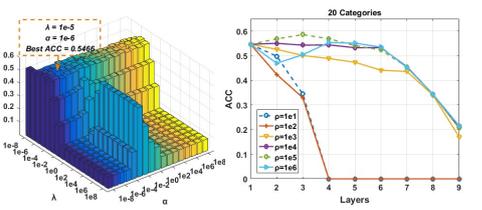

Figure 8 (b). 20 categories from UMIST.

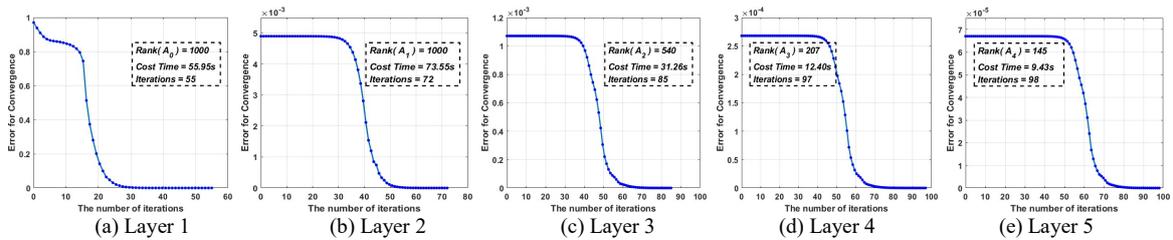

Figure 9. Convergence analysis of DeepLRR algorithm on the COIL100 object database.